\documentclass[letterpaper]{article} % DO NOT CHANGE THIS
\usepackage{aaai2026}  % DO NOT CHANGE THIS
\usepackage{times}  % DO NOT CHANGE THIS
\usepackage{helvet}  % DO NOT CHANGE THIS
\usepackage{courier}  % DO NOT CHANGE THIS
\usepackage[hyphens]{url}  % DO NOT CHANGE THIS
\usepackage{graphicx} % DO NOT CHANGE THIS
\usepackage{natbib}  % DO NOT CHANGE THIS AND DO NOT ADD ANY OPTIONS TO IT
\usepackage{caption} % DO NOT CHANGE THIS AND DO NOT ADD ANY OPTIONS TO IT
\usepackage{algorithm}
\usepackage{algorithmic}

\usepackage{newfloat}
\usepackage{listings}

\usepackage{booktabs}       % professional-quality tables
\usepackage{amsfonts}       % blackboard math symbols
\usepackage{nicefrac}       % compact symbols for 1/2, etc.
\usepackage{microtype}      % microtypography
\usepackage{xcolor}         % colors
\usepackage{amsmath} %zzx
\usepackage{booktabs}
\usepackage{multirow}
\usepackage[table]{xcolor}
\usepackage{colortbl}
\usepackage[dvipsnames]{xcolor}
\usepackage{amssymb} % for \checkmark
\newcommand{\tablestyle}[2]{\setlength{\tabcolsep}{#1}\renewcommand{\arraystretch}{#2}\centering\footnotesize}
\def\Alg{SpecQuant }
\DeclareCaptionStyle{ruled}{labelfont=normalfont,labelsep=colon,strut=off} % DO NOT CHANGE THIS
\floatstyle{ruled}
\newfloat{listing}{tb}{lst}{}
\floatname{listing}{Listing}
\title{SpecQuant: Spectral Decomposition and Adaptive Truncation for Ultra-Low-Bit LLMs Quantization}
\author{
    Zhixiong Zhao\textsuperscript{\rm 1,3}\equalcontrib\thanks{This work was done when Zhixiong Zhao was an intern at Shanghai Jiao Tong University. Fangxin Liu and Li Jiang are the corresponding authors.}, Fangxin Liu\textsuperscript{\rm 1,2}\equalcontrib, Junjie Wang\textsuperscript{\rm 1,2}, Chenyang Guan\textsuperscript{\rm 1}, Zongwu Wang\textsuperscript{\rm 1,2}\\ Li Jiang\textsuperscript{\rm 1,2}, Haibing Guan\textsuperscript{\rm 1}
}
\affiliations{
    \textsuperscript{\rm 1}Shanghai Jiao Tong University,\\
    \textsuperscript{\rm 2}Shanghai Qi Zhi Institute, \\
    \textsuperscript{\rm 3}Nanyang Technological University\\
    zhixiong003@e.ntu.edu.sg, \,\, \{liufangxin, ljiang\_cs\}@sjtu.edu.cn
}

\begin{document}
\maketitle
\begin{abstract}
The emergence of accurate open large language models (LLMs) has sparked a push for advanced quantization techniques to enable efficient deployment on end-user devices. In this paper, we revisit the challenge of extreme LLM compression---targeting ultra-low-bit quantization for both activations and weights---from a Fourier frequency domain perspective.
We propose \Alg, a two-stage framework that tackles activation outliers and cross-channel variance. In the first stage, activation outliers are smoothed and transferred into the weight matrix to simplify downstream quantization. In the second stage, we apply channel-wise low-frequency Fourier truncation to suppress high-frequency components while preserving essential signal energy, improving quantization robustness. Our method builds on the principle that most of the weight energy is concentrated in low-frequency components, which can be retained with minimal impact on model accuracy. To enable runtime adaptability, we introduce a lightweight truncation module during inference that adjusts truncation thresholds based on channel characteristics. On LLaMA-3 8B, \Alg achieves 4-bit quantization for both weights and activations, narrowing the zero-shot accuracy gap to only 1.5\% compared to full precision, while delivering 2× faster inference and 3× lower memory usage. Code will be available at \url{https://github.com/Kishon-zzx/SpecQuant}.
\end{abstract}

\section{Introduction}
Large Language Models (LLMs)~\citep{touvron2023llama,guo2025deepseek,achiam2023gpt} have demonstrated impressive performance across a wide range of natural language processing tasks, including code generation and open-ended reasoning. These advances are largely driven by massive model scales and extensive pretraining data. However, the resulting memory footprint and computational cost pose significant challenges for deployment on edge devices~\citep{lin2024duquant,heo2023rethinking}.

To reduce memory and accelerate inference, recent quantization methods~\citep{liu2024spinquant,hu2025ostquant} aim to represent both weights and activations in low-bit formats~\citep{zhao2025quarkquantizationenabledcircuitsharing,lin2024awq}, enabling faster matrix multiplications and smaller model sizes~\citep{xiao2023smoothquant}. Despite their effectiveness, a core challenge remains: \textit{activation outliers}, which expand the dynamic range and induce significant accuracy degradation when quantized~\citep{wei2022outlier}.

\begin{figure}
    \centering
    \includegraphics[width = 0.48\textwidth]{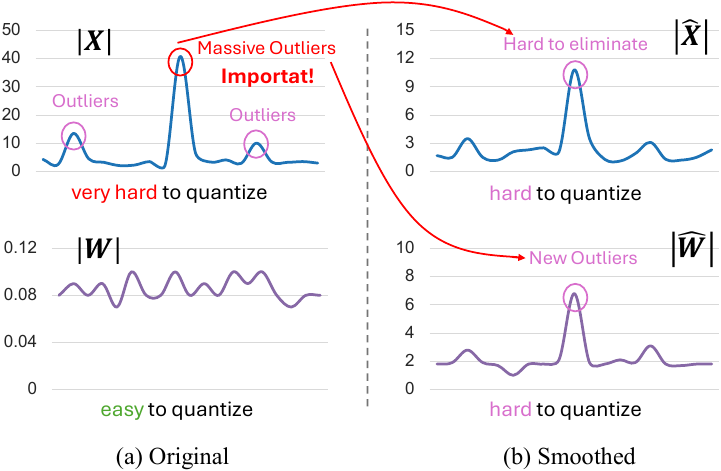}
    \caption{Activation and weight distributions before and after naive smoothing. While smoothing aims to mitigate activation outliers, it often transfers the quantization burden to weights, introducing new outliers and degrading the robustness of both activations and weights under quantization.}
    \label{fig:motivation}
    %\vspace{-0.6cm}
\end{figure}

Existing approaches attempt to mitigate outlier impact via distributional transformations. SmoothQuant~\citep{xiao2023smoothquant} shifts activation outliers into weights via layer-wise scaling; OSTQuant~\citep{hu2025ostquantrefininglargelanguage}, SpinQuant~\citep{liu2024spinquant} and QuaRot~\citep{ashkboos2024quarot} employ lightweight rotation layers to reduce activation variance; SVDQuant~\citep{li2024svdqunat} applies global low-rank approximation to absorb outliers. However, these strategies face fundamental limitations. As shown in Figure~\ref{fig:motivation}, smoothing-based methods often transfer the burden of quantization from activations to weights without eliminating it. Rotation-based techniques introduce non-negligible runtime overhead, while SVD-based approximations fail to preserve channel-wise outlier structures critical for contextual understanding.

Recent work~\citep{jin2025massivevaluesselfattentionmodules} further reveals that extreme activation values often encode fine-grained contextual cues, essential for reasoning tasks. Thus, indiscriminate quantization of these values leads to substantial performance loss on long-context benchmarks. This motivates the need for a more principled and robust quantization strategy that preserves informative outliers without incurring high computational cost.

In this work, we propose SpecQuant, a novel compression framework based on \textbf{adaptive Fourier-domain decomposition}, which explicitly targets the spectral structure of weights induced by smoothed activations. \Alg consists of two stages: (1) activation smoothing to migrate outliers into the weight domain, and (2) channel-wise low-frequency truncation to suppress the transferred high-frequency noise while preserving signal fidelity. We observe that weights exhibit strong low-frequency bias in the Fourier domain. This property allows us to truncate high-frequency components with minimal impact on model accuracy, enabling more robust low-bit quantization:
\noindent\textbf{Our contributions are summarized as follows:}
\begin{itemize}
    \item \textbf{Frequency Domain Approximation:} We are the first to bridge a connection between frequency-domain compression and quantization robustness in LLMs. Our analysis leverages Fourier energy decay properties to provide theoretical guarantees for preserving accuracy under aggressive quantization.

    \item \textbf{Outlier-Resilient Spectral Quantization:} We propose \Alg, a novel two-stage quantization framework that first absorbs activation outliers via scaling-based smoothing and then performs adaptive, channel-wise spectral truncation in the Fourier domain, effectively mitigating quantization error caused by outlier redistribution.
    
    \item \textbf{Extensive Evaluation:} We evaluate \Alg on eight LLMs across ten datasets, showing up to 3× memory reduction and 1.7× speedup with only 1.5\% accuracy drop, outperforming prior SOTA methods.
\end{itemize}

\section{Related Works}
\subsection{LLM Compression and Outlier Mitigation}

Recent progress in post-training quantization of LLMs below 4-bit precision---such as SmoothQuant \cite{xiao2023smoothquant}, OPTQ \cite{frantar2023optq}, QuIP \cite{chee2023quip}, SpinQuant \cite{liu2024spinquant}, QuaRot \cite{ashkboos2024quarot}, SVDQuant \cite{li2024svdqunat}, and QuIP\# \cite{tseng2024quip}---has focused on mitigating activation and weight outliers. These outliers, particularly extreme weight values, expand the quantization dynamic range, leading to significant accuracy degradation.
SmoothQuant addresses this by shifting quantization difficulty from activations to weights via layer-wise scaling. SpinQuant and QuaRot introduce rotation-based strategies: SpinQuant learns orthogonal transforms to align weight-activation distributions, while QuaRot applies random rotations to suppress outliers. SVDQuant builds on SmoothQuant by decomposing weight matrices using singular value decomposition (SVD), isolating compressible residuals.
Meanwhile, QuIP and QuIP\# leverage randomized matrix transformations and E8 lattice structures for improved vector quantization. Both adopt error-compensated column-wise quantization from OPTQ, propagating residuals to enhance later quantization steps.

Despite these advances, challenges remain. Migration-based methods struggle with extreme or dense outlier distributions, rotation-based techniques introduce inference overhead, and global decomposition methods often fail to capture channel-specific outlier patterns. These limitations hinder robust ultra-low-bit quantization for real-world LLM deployment.

\subsection{Frequency-Domain Optimization Strategies}
Frequency-domain optimization has emerged as a powerful tool across machine learning, offering unique advantages in compression and efficiency. In computer vision, spectral representations help stabilize training and improve generalization. For example, FcaNet \cite{qin2021fcanet} applies discrete cosine transforms (DCT) to enhance channel attention, while F3D \cite{liu2021f3d} replaces 3D convolutions with frequency-domain operations to improve hardware efficiency. In time-series forecasting, FreDF \cite{wang2024fredf} uses spectral decomposition with consistency constraints to address temporal error propagation.
Recently, frequency-domain principles have begun influencing LLM optimization. FourierFT \cite{gao2024parameter} enables parameter-efficient fine-tuning by learning sparse frequency coefficients and reconstructing full model weights via inverse DFT. Notably, frequency projections align well with LLM channel structures and naturally absorb multi-scale activation outliers through bandwidth filtering.

This compatibility motivates the use of frequency-domain decomposition as an effective strategy for robust quantization. However, despite this promising synergy, systematic exploration of frequency-domain quantization for LLMs remains limited in current efforts.

\begin{figure*}[ht]
    \centering
    \includegraphics[width=\linewidth]{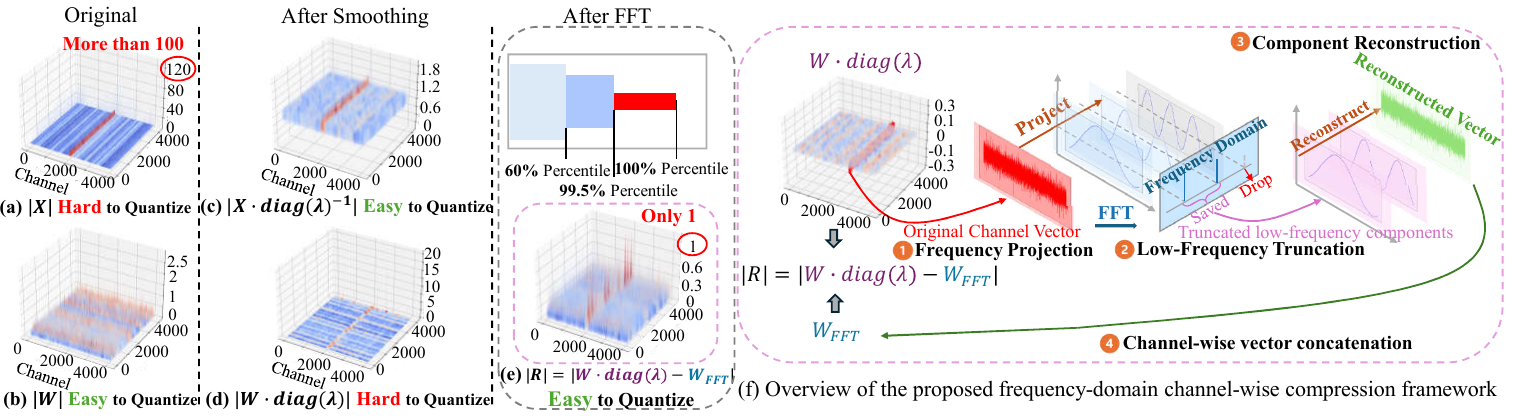}
    \caption{ Overview of the proposed \Alg.}
    \label{fig:overview}
    %\vspace{-0.6cm}
\end{figure*}

\section{Method}
In this section, we first establish the theoretical foundation of Low-Frequency Fourier Projection for channel vectors, laying the groundwork for our subsequent methodology. We then introduce \Alg, a novel quantization paradigm specifically designed for LLMs. The core innovation of our approach is an auxiliary low-frequency spectral truncation branch that effectively mitigates channel-wise quantization challenges in both weight matrices and activation tensors, as illustrated in Figure~\ref{fig:overview}.

\subsection{Preliminaries}
\textbf{Quantization} 
Multi-head Self-Attention (MSA) and Feed-Forward Network (FFN), the core components within each Transformer block of LLMs, fundamentally rely on standard linear transformations formulated as  $Y = X \cdot W \in \mathbb{R}^{T \times C_{\text{out}}}$, where, $X \in \mathbb{R}^{T \times C_{\text{in}}}$ denotes the input activation matrix and $W \in \mathbb{R}^{C_{\text{in}} \times C_{\text{out}}}$ is the corresponding weight matrix. In this work, we apply uniform integer quantization to both activations and weights to improve hardware efficiency. Specifically, a $b$-bit quantization maps a floating-point tensor $X$ into a low-bit integer representation $X_q$ as follows:
\begin{equation}
X_q = \left\lfloor \text{clamp} \left( \frac{X}{\Delta} + z,\, 0,\, 2^b - 1 \right) \right\rceil
\end{equation}
where $\Delta = \frac{\max(X) - \min(X)}{2^b - 1}$ is the quantization step size, and $z = -\frac{\min(X)}{\Delta}$ is the zero-point offset. The operator $\lfloor \cdot \rceil$ denotes rounding to the nearest integer.
Following prior works~\citep{frantar2022gptq,lin2024duquant}, we adopt per-token quantization for activations and per-channel quantization for weights. 

\noindent\textbf{Frequency Domain Projection}
Within each Transformer block, the linear transformation $Y = X \cdot W \in \mathbb{R}^{T \times C_{\text{out}}}$ remains the fundamental operation, where $X \in \mathbb{R}^{T \times C_{\text{in}}}$ and $W \in \mathbb{R}^{C_{\text{in}} \times C_{\text{out}}}$  are as defined above. To facilitate compression of weight matrices while maintaining hardware compatibility, we propose projecting weight matrices into the frequency domain using the Discrete Fourier Transform (DFT):
\begin{equation}
\resizebox{1\hsize}{!}{$
W_{\text{freq}}[k] = \sum_{n=0}^{N-1} W[n] \cdot e^{-i \frac{2\pi k n}{N}}, \quad k = 0, 1, \dots, N - 1
$}
\end{equation}
where \( N \) denotes the size of the input dimension. Due to the  \( \mathcal{O}(N^2) \) complexity of naive DFT computation, we leverage the Fast Fourier Transform (FFT), which reduces this to \( \mathcal{O}(N \log N) \), and is efficiently implemented in libraries such as CUDA FFT. The FFT can be recursively defined as:
\begin{equation}
\resizebox{\hsize}{!}{$
W_{\text{freq}} = \text{FFT}(W) = \text{Combine}\left( \text{FFT}(W_{\text{even}}), \text{FFT}(W_{\text{odd}}) \right)
$}
\end{equation}

By preserving low-frequency components, which concentrate the majority of the signal energy, and discarding less informative high-frequency components, this method achieves significant compression. Also, it maintains hardware-agnostic deployment flexibility, making it a practical and scalable solution for efficient LLM inference.

\subsection{Low-Frequency Fourier Projection of Channel Vectors}

Existing studies have confirmed that the outlier characteristics of LLM weights and activations exhibit significant cross-channel heterogeneity—dynamic range differences between channels can reach 3-4 orders of magnitude, and the proportion of high-frequency energy shows a strong negative correlation with task relevance \citep{lin2024duquant,liu2024spinquant}. To precisely capture such channel-specific properties, we propose a \textbf{channel-independent frequency-domain projection framework}, with the core assumption that each output channel's weight vector $\mathbf{W}[:,j] \in \mathbb{R}^{C_{\text{in}}}$ in LLMs can be treated as an independent stationary signal, whose energy is primarily concentrated in low-frequency components. This assumption is empirically supported: in the attention layers of LLaMA-2 7B, the average low-frequency (top 20\% frequencies) energy proportion of 1000 randomly sampled channel vectors reaches 92.3\%, with a standard deviation of only 3.7\%.

For the weight vector $\mathbf{W}[:,j] = [w_0, w_1, \dots, w_{C_{\text{in}}-1}]^\top \in \mathbb{R}^{C_{\text{in}}}$ of the $j$-th output channel (where $w_n \in \mathbb{R}$ are real-valued weights), its discrete Fourier transform is defined as:
\begin{equation}
\mathbf{W}_{\text{freq}}[k, j] = \sum_{n=0}^{C_{\text{in}}-1} w_n \cdot e^{-i \frac{2\pi k n}{C_{\text{in}}}}, \quad k = 0, 1, \dots, C_{\text{in}} - 1 
\end{equation}
where $i$ is the imaginary unit, $k$ is the frequency index, and $e^{-i \frac{2\pi k n}{C_{\text{in}}}} = \cos\left(\frac{2\pi k n}{C_{\text{in}}}\right) - i\sin\left(\frac{2\pi k n}{C_{\text{in}}}\right)$ is the complex exponential function.
Since $\mathbf{W}[:,j]$ is a real-valued vector, its DFT coefficients satisfy conjugate symmetry:
\begin{equation}
    \mathbf{W}_{\text{freq}}[k, j] = \overline{\mathbf{W}_{\text{freq}}[C_{\text{in}}-k, j]}
\end{equation} 
where $\overline{\cdot}$ denotes complex conjugation. This property allows us to store only the first $\lceil C_{\text{in}}/2 \rceil$ coefficients for complete reconstruction of the original vector, naturally reducing the frequency-domain storage overhead by 50\%.

Specifically, each channel vector $\mathbf{W}[:, j] \in \mathbb{R}^{C_{\text{in}}}$ is independently transformed into the frequency domain via FFT:
\begin{equation}
\begin{aligned}
\mathbf{W}_{\text{freq}}[:, j] = \texttt{FFT}(\mathbf{W}[:, j]) = \sum_{n=0}^{C_{\text{in}}-1} \mathbf{W}[n, j] \cdot e^{-i 2\pi k n / C_{\text{in}}}, \\\quad k \in \{0, 1, \ldots, C_{\text{in}}-1\}
\end{aligned}
\end{equation}
This yields complex spectral coefficients $X_k \in \mathbb{C}$ with $O(C_{\text{in}} \log C_{\text{in}})$ complexity. By applying strategic spectral sparsification—contrasting with global SVD-based approximations—\Alg achieves fine-grained outlier suppression at channel level.

Similar to how SVD retains top-$k$ singular values for approximation, Fourier analysis shows that smooth signals concentrate most energy in low-frequency components. This provides a theoretical basis for our compression approach. Formally, given a real-valued discrete signal $x[n]$ of length $N$ with FFT:
\begin{equation}
\resizebox{\linewidth}{!}{$
X[k] = \sum_{n=0}^{N-1} x[n] \cdot e^{-i 2\pi kn / N}, \quad k = 0, 1, \ldots, N-1.
$}
\end{equation}
Parseval’s theorem \citep{4180434} ensures energy preservation between time and frequency domains:
\begin{equation}
\sum_{n=0}^{N-1} |x[n]|^2 = \frac{1}{N} \sum_{k=0}^{N-1} |X[k]|^2.
\end{equation}
Here, the squared magnitude $|X[k]|^2$ represents the energy contribution at frequency $k$.

Assuming $x[n]$ samples a continuous function $f(t)$ with $r$ continuous derivatives ($f \in C^r$), classical Fourier theory states that smoother functions have faster decaying Fourier coefficients:
\begin{equation} 
\label{eq}
|X[k]| \leq \frac{C}{|k|^r}, \quad \text{for some constant } C,
\end{equation}
As $k$ increases, $|X[k]|$ rapidly decreases, making high-frequency energy negligible. Thus, most energy concentrates in low frequencies, justifying truncated Fourier representations to approximate the original signal with minimal information loss.

Based on this, Figure~\ref{fig:overview} (f) illustrates the compression and reconstruction of a channel vector $\mathbf{W}[:, j] \in \mathbb{R}^{C_{\text{in}}}$ through four steps:

\textbf{1) Frequency Projection:} Transform the channel vector to the frequency domain using FFT, yielding $N=C_{\text{in}}$ frequency components characterized by magnitude $A_k$, phase $\phi_k$, and frequency $f_k$.

\textbf{2) Low-Frequency Truncation:}
Given a target compression ratio $\rho$, select $k = \lfloor \rho \cdot N / 3 \rfloor$ low-frequency components, since each stores three parameters.

\textbf{3) Component Reconstruction:}
Reconstruct each retained component in time domain as $H_m[n] = A_m \cdot e^{i(2\pi f_m n + \phi_m)}$, for $n = 0, 1, \dots, N - 1$.

\textbf{4) Channel-wise vector concatenation:} Concatenate reconstructed vectors channel-wise to form the compressed representation, preserving structure for downstream processing and quantization.
% After reconstructing the individual truncated components in the time domain, the resulting channel-wise vectors are concatenated to form the final compressed representation. This step preserves the structural integrity of the channel dimensions and enables efficient downstream processing. As shown in Figure \ref{fig:overview}, this concatenated vector is then integrated back into the quantization pipeline.

To further reduce memory, we exploit the relation $f_k = k/N$, allowing implicit frequency indexing without storing $f_k$. Thus, each component stores only $(A_k, \phi_k)$, reducing memory by 33\%. The number of retained components updates to $k = \lfloor \rho \cdot N / 2 \rfloor$, with reconstruction:

\begin{equation}
\resizebox{\linewidth}{!}{$
H_m[n] = A_m \cdot e^{i\left( \frac{2\pi m}{N} n + \phi_m \right)}, \quad n \in \{0, 1, \ldots, N-1\}
$}
\end{equation}
This implicit frequency indexing strategy ensures no additional approximation error, as spectral positions remain unchanged under channel-wise FFT.

\textbf{Mathematical Guarantee:} Let $\mathcal{F}^{-1}$ be the inverse FFT. The reconstruction error $\epsilon$ satisfies:
\begin{equation}
\resizebox{\linewidth}{!}{$
\epsilon = \left\| \mathbf{W}[:, j] - \mathcal{F}^{-1}\left( \sum_{m=0}^{k-1} H_m \right) \right\|_2 
\leq \sqrt{ \sum_{m=k}^{N-1} |A_m|^2 }
$}
\end{equation}
As per the decay property above, this upper bound decreases rapidly for smooth vectors, allowing high compression with minimal fidelity loss.
% This upper bound decays rapidly for smooth vectors, as discussed in Eq.~(\ref{eq}), ensuring that high compression rates can be achieved without sacrificing reconstruction fidelity.

\begin{figure*}[h!]
    \centering
    \includegraphics[width=\linewidth]{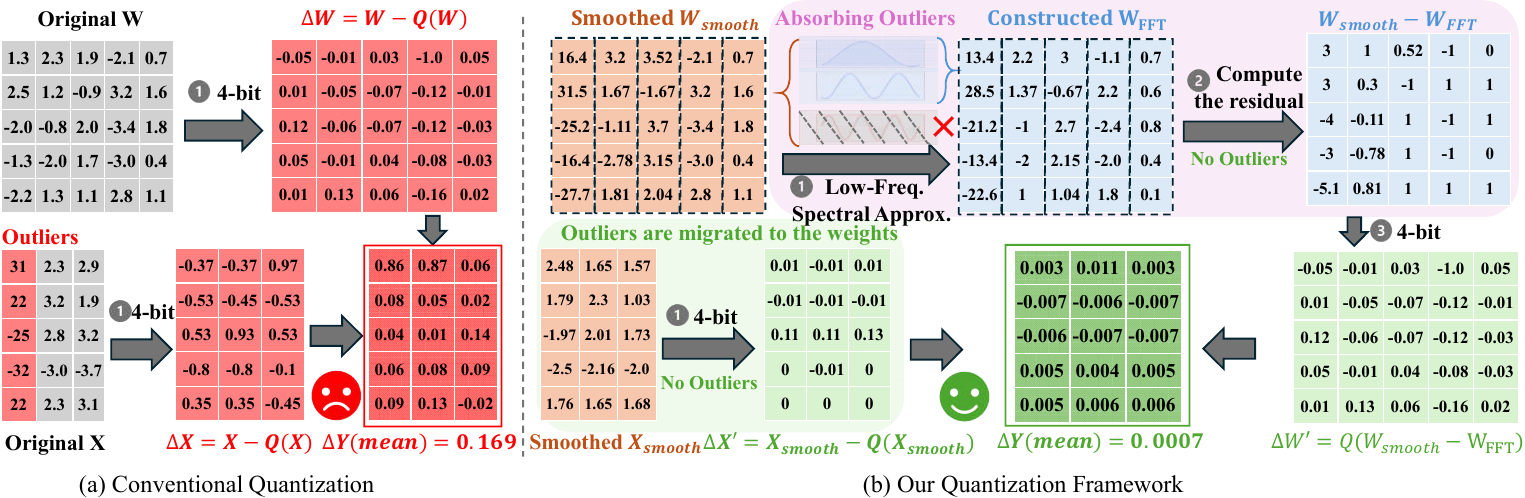}
    \caption{Comparison between conventional quantization and \Alg. Outlier channels in input activations are marked in \textcolor{red}{red}. \Alg adaptively absorbs these outliers using frequency-domain approximation, reducing overall quantization error.}
    % \caption{ Comparison between Conventional Quantization Method and \Alg. Outlier channels in input activations (X) are highlight in red, and \Alg takes these into consideration, which can contribute to a reduction in quantization error.}
    \label{fig:visiual}
    %\vspace{-0.6cm}
\end{figure*}

\subsection{\Alg Framework}
\textbf{Migrate outliers from activation to weight.}
To mitigate activation outliers, recent methods~\citep{xiao2023smoothquant,lin2024awq} propose jointly scaling activations and weights. Specifically, the input $\mathbf{X}$ is scaled by a per-channel factor $\boldsymbol{\lambda} \in \mathbb{R}^m$, yielding smoothed activations
$\hat{\mathbf{X}} = \mathbf{X} \cdot \text{diag}(\boldsymbol{\lambda})^{-1}$,
which suppresses extreme values and reduces activation quantization error (Figure~\ref{fig:overview} (a) and (c)). To preserve the original computation, the weight matrix is adjusted as
$\hat{\mathbf{W}} = \mathbf{W} \cdot \text{diag}(\boldsymbol{\lambda})$.
% which exhibits reduced magnitude and fewer extreme values, effectively lowering the activation quantization error, as shown in Figure~\ref{fig:overview}(a)(c). To maintain functional equivalence, the weight matrix is transformed as
% $\hat{\mathbf{W}} = \mathbf{W} \cdot \text{diag}(\boldsymbol{\lambda})$,
However, this increases the dynamic range and magnitude of the weight values (Figure~\ref{fig:overview} (b) and (d)), introducing new quantization challenges. This trade-off limits the net benefit of such smoothing-based approaches.

\noindent\textbf{Absorbing Weight Outliers via Channel-wise Low-Frequency Spectral Approximation.}
To address the increased weight quantization difficulty after smoothing, we propose an activation-aware channel-wise spectral compression framework that absorbs outliers by retaining low-frequency weight components in the Fourier domain, while dynamically allocating frequency budget based on actual activation-weight interaction strength. Unlike traditional low-rank approximations such as SVD, our method preserves channel-wise structure, leverages intrinsic weight smoothness, and prioritizes components based on their real impact on model outputs.

Our key insight is twofold: (1) Weight outliers often manifest as sharp local variations corresponding to high-frequency components in the frequency domain; (2) Instead of relying solely on weight characteristics, channel importance should be determined by their interaction with input activations—since this interaction ultimately drives model outputs. By allocating more frequency budget to channels with more important channels, we adaptively preserve critical components while suppressing noise-like outliers.

For each channel $\hat{\mathbf{W}}[:, j] \in \mathbb{R}^{C_{\text{in}}}$, we first compute an activation-aware importance score that quantifies its contribution strength in activation-weight interactions:
\begin{equation}
\text{Score}(j) = \left| \bar{\mathbf{X}}_{:,j} \cdot \bar{\hat{\mathbf{W}}}_{:,j} \right|
\end{equation}
where $\bar{\mathbf{X}}_{:,j} \in \mathbb{R}$ denotes the average value of the $j$-th channel in the input matrix, $\bar{\hat{\mathbf{W}}}_{:,j} \in \mathbb{R}$ represents the average value of the $j$-th channel in the weight matrix, and $\cdot$ indicates scalar multiplication. This score captures how significantly the channel influences outputs.

We then apply FFT to transform the channel vector into the frequency domain:
\begin{equation}
\hat{\mathbf{W}}_{\text{freq}}[:, j] = \mathcal{F}(\hat{\mathbf{W}}[:, j])
\end{equation}
A softmax-based normalization over the activation-aware scores determines the frequency budget allocation:
\begin{equation}
\rho_j = \frac{\exp(\alpha \cdot \text{Score}(j))}{\sum_{l=1}^{C_{\text{out}}} \exp(\alpha \cdot \text{Score}(l))},
\end{equation}
where $\alpha$ is a tunable temperature parameter. Each channel retains $k_j = \lfloor \rho_j \cdot C_{\text{in}} \rfloor$ lowest-frequency components, with channels having higher activation impact receiving more budget to preserve critical spectral information.

The compressed weights are reconstructed via inverse FFT:
\begin{equation}
\mathbf{W}'[:, j] = \mathcal{F}^{-1}(\text{Truncate}(\hat{\mathbf{W}}_{\text{freq}}[:, j]))
\end{equation}
This activation-aware strategy dynamically optimizes frequency retention at the channel level, ensuring that components most influential to actual model behavior are prioritized—resulting in a fine-grained compression scheme that balances efficiency and performance.

The residual $\mathbf{R} = \hat{\mathbf{W}} - \mathbf{W}'$ is quantized separately. The overall matrix product is approximated as:
\begin{equation}
\begin{gathered}
\mathbf{X} \mathbf{W} = \hat{\mathbf{X}} \hat{\mathbf{W}} = 
\hat{\mathbf{X}} \mathbf{W}' + \hat{\mathbf{X}} \mathbf{R} \\
\approx 
\underbrace{\hat{\mathbf{X}} \mathbf{W}'}_{\text{16-bit low-frequency branch}} + 
\underbrace{Q(\hat{\mathbf{X}}) Q(\mathbf{R})}_{\text{4-bit residual}}
\end{gathered}
\end{equation}
This formulation retains the dominant low-frequency components in higher precision, while compressing the residual with low-bit quantization. Empirically, setting $k$ to 16 or 32 per channel ensures both compression and accuracy. The overhead is negligible, adding only $2k/m$ to the overall cost, where $m$ is the number of input channels.

% Here, $\mathbf{W}'$ is retained in higher precision (e.g., 16-bit), while the quantization-friendly residual $\mathbf{R}$ is compressed with low-bit precision.

% From the perspective of signal analysis, this approach is theoretically grounded by Parseval’s theorem, which guarantees that the total signal energy is preserved in the frequency domain:
From the spectral perspective, our method is inspired by Parseval’s theorem, which ensures the energy of a signal is preserved in the frequency domain:
\begin{equation}
\sum_{n=0}^{N-1} |x[n]|^2 = \frac{1}{N} \sum_{k=0}^{N-1} |X[k]|^2
\end{equation}
Moreover, for smooth functions $f \in C^r$, their Fourier coefficients decay polynomially as:
\begin{equation}
|X[k]| \leq \frac{C}{|k|^r}
\end{equation}
which indicating that most informative structure lies in low frequencies, justifying our truncation strategy for quantization robustness.

As illustrated in Figure~\ref{fig:visiual}(a), migrating activation outliers into weights without addressing the amplified weight magnitude leads to quantization artifacts, especially when activations exhibit large variance. This explains the limited effectiveness of conventional smoothing methods in compressing LLMs. In contrast, Figure~\ref{fig:visiual}(b) shows that \Alg absorbs migrated outliers through spectral approximation, yielding a quantization-friendly residual matrix $\Delta W'$. This design ensures that both smoothed activations and adapted weights are effectively quantized, resulting in improved end-to-end compression quality. 
We provide the full algorithmic description of our framework in Appendix.~\ref{sec:A}.

% Even though the dominant cause of variation in $\Delta Y$ originates from large-magnitude activations (rather than weight perturbations), it can still induce considerable output deviations in the final model. As a result, merely quantizing the weights alone leads to noticeable performance degradation.

% In contrast, as shown in Figure~\ref{fig:visiual}(b), \Alg effectively absorbs the outliers that have been migrated from $X$, producing a quantization-friendly residual matrix $\Delta W'$. Empirically, the number of preserved low-frequency components $k$ is typically set to 16 or 32. Therefore, the additional parameters and computations introduced by this spectral branch are negligible, contributing only $2k/m$ to the overall cost, where $m$ denotes the number of input channels in the original matrix. We present the algorithm of the entire compression framework in Appendix \ref{sec:A}.
\begin{table*}[ht]
\centering
\caption{Comparison of perplexity on WikiText2 and averaged accuracy on nine Zero-Shot tasks. Results for SmoothQuant, GPTQ, OmniQuant, AWQ, and QuaRot are based on official code and SpinQuant’s results for LLaMA-2/3 using official weights, with LLaMA-1 from the official code.}
%\vspace{-0.3cm}
\tablestyle{1pt}{1.3}
\resizebox{\textwidth}{!}{ % Resize to fit within the text width
\begin{tabular}{|c|c|cc|cc|cc|cc|cc|cc|cc|cc|}
\hline
\multirow{2}{*}{\#Bits}  &\multirow{3}{*}{Method} & \multicolumn{2}{c}{LLaMA-3 8B}  & \multicolumn{2}{c}{LLaMA-3 70B} &  \multicolumn{2}{c}{LLaMA-2 7B} & \multicolumn{2}{c}{LLaMA-2 13B} & \multicolumn{2}{c}{LLaMA-2 70B} & \multicolumn{2}{c}{LLaMA 7B} & \multicolumn{2}{c}{LLaMA 13B} &\multicolumn{2}{c}{LLaMA 30B}\\
 &  &0-shot$^9$ &Wiki &0-shot$^9$ &Wiki&0-shot$^9$ &Wiki&0-shot$^9$ &Wiki&0-shot$^9$ &Wiki&0-shot$^9$ &Wiki&0-shot$^9$ &Wiki&0-shot$^9$ &Wiki \\
W-A-KV & &Avg.($\uparrow$) &PPL($\downarrow$) &Avg.($\uparrow$) &PPL($\downarrow$)&Avg.($\uparrow$) &PPL($\downarrow$)&Avg.($\uparrow$) &PPL($\downarrow$)&Avg.($\uparrow$) &PPL($\downarrow$)&Avg.($\uparrow$) &PPL($\downarrow$)&Avg.($\uparrow$) &PPL($\downarrow$)&Avg.($\uparrow$) &PPL($\downarrow$) \\
\hline
16-16-16  & FP16 & 68.09 & 6.14 & 73.81 & 2.86 & 65.21 & 5.47 & 67.61 & 4.88 & 71.59 & 3.32 &64.48 & 5.68 &66.67 &5.09 &70.00 &4.10 \\
\hline
\multirow{8}{*}{4-16-16}  & RTN & 63.70 & 8.13 & 31.15 & 1e5 & 61.27 & 7.02 & 60.24 & 6.39 & 69.62 & 3.87 & 62.67 & 7.94 &63.45 &8.60 &65.69 &6.13 \\
  & SmoothQuant & 62.79 & 8.12 & 67.94 & 6.70 & 58.88 & 8.03 & 62.03 & 5.86 & 65.93 & 5.50 & 62.24 & 7.46 &62.69 &18.75 &65.69 &5.80 \\
 & GPTQ & 61.03 & 7.43 & 31.45 & 9e3 & 60.86 & 9.84 & 64.71 & 5.79 & 70.96 & 3.94 & 60.15 & 7.93 &64.36 &6.58 &66.95 &5.26 \\
  & Omniquant & 65.66 & 7.19 & - & - & 63.19 & 5.74 & 66.38 & 5.02 & 71.04 & 3.47 & 63.42 & 5.86 &66.22 &5.21 &69.07 &4.25 \\
  & AWQ & 67.03 & 7.36 & 68.92 & 5.92 & 63.89 & 5.83 & 66.25 & 5.07 & 70.88 & 4.03 & 63.30 & 5.97 &65.58 &5.28 &69.44 &4.28 \\
  & QuaRot & 67.27 & 6.53 & 72.93 & 3.53 & 64.30 & 5.62 & 66.95 & 5.00 & 71.21 & 3.41 & 63.40 & 5.83 &65.91 &5.20 &69.73 &4.27 \\
  & SpinQuant & 66.54 & 6.49 & 72.90 & \textbf{3.49} & 63.59 & 5.58 & \textbf{67.14} & 5.00 & 71.12 & 3.43 & 63.94 & \textbf{5.76} &66.32 &\textbf{5.16} &69.62 &4.21 \\
   \rowcolor{green!10}
  & \textbf{\Alg} &\textbf{66.88}  &\textbf{6.48}  &\textbf{72.98}  &3.53  &\textbf{63.99}  &\textbf{5.56}  &67.10  &\textbf{4.98}  &\textbf{71.12}  &\textbf{3.40}  &\textbf{63.99}  &5.85  &\textbf{66.35} &5.21 &\textbf{69.77} &\textbf{4.21}\\
\hline
\multirow{6}{*}{4-4-16} & RTN & 33.42 & 6e2 & 31.21 & 8e3 & 32.44 & nan & 30.86 & 8e3 & 30.90 & 7e4 & 32.51 & 7e3 &31.63 &3e4 &31.57 &2e3 \\
  & SmoothQuant & 33.04 & 1e3 & 34.67 & 2e2 & 32.13 & nan & 34.26 & 1e3 & 35.86 & 3e2 & 34.42 & 3e2 &33.29 &6e2 &34.64 &1e3 \\
  & GPTQ & 32.98 & 5e2 & 31.47 & 4e4 & 32.72 & nan & 30.11 & 4e3 & 30.86 & nan & 32.12 & 1e3 &31.51 &3e3 &30.88 &2e3 \\
  & QuaRot & 61.69 & 8.02 & 65.56 & 6.35 & 61.87 & 6.05 & \textbf{65.13} & 5.35 & 69.96 & 3.78 & 61.76 & 6.22 &64.46 &5.50 &68.14 &4.57 \\
  & SpinQuant & 64.11 & 7.28 & 66.99 & 6.10 & 57.37 & 6.78 & 63.23 & 5.24 & 70.58 & 3.68 & 61.82 & 6.08 &64.59 &\textbf{5.36} &68.08 &4.53 \\
   \rowcolor{green!10}
  & \textbf{\Alg} &\textbf{64.75}  &\textbf{7.25}  &\textbf{69.75}  &\textbf{5.12}  &\textbf{62.88}  &\textbf{5.88}  &65.12  &\textbf{5.18}  &\textbf{70.77}  &\textbf{3.63}  &\textbf{61.85}  &\textbf{6.05} &\textbf{64.77} &5.45 &\textbf{68.25} &\textbf{4.50}  \\
\hline
\multirow{7}{*}{4-4-4}  & RTN & 33.18 & 7e2 & 30.82 & 8e3 & 32.67 & nan & 30.93 & 7e3 & 31.73 & 7e4 & 32.87 & 1e4 &31.33 &3e4 &31.64 &2e3 \\
  & SmoothQuant & 32.96 & 1e3 & 33.76 & 3e2 & 32.12 & nan & 33.36 & 1e3 & 35.54 & 3e2 & 34.42 & 3e2 &33.28 &5e2 &34.65 &1e3\\
  & GPTQ & 33.71 & 6e2 & 31.20 & 4e4 & 33.52 & nan & 27.85 & 5e3 & 31.09 & nan & 31.80 & 2e3 &30.63 &3e3 &31.07 &2e3\\
 & Omniquant & 32.33 & 4e2 & - & - & 48.40 & 14.26 & 50.35 & 12.30 & - & - & 48.46 & 11.26 &45.63 &10.87 &45.04 &12.35 \\
  & QuaRot & 61.38 & 8.18 & 65.33 & 6.6 & 61.48 & 6.11 & 65.16 & 5.39 & 70.30 & 3.80 & 61.22 & 6.26 &64.59 &5.53 &68.08 &4.60 \\
  & SpinQuant & 64.10 & 7.35 & 66.31 & 6.24 & 62.01 & 5.96 & 64.13 & 5.74 & 70.57 & 3.61 & 61.32 & 6.12 &64.95 &5.39 &\textbf{68.14} &4.55 \\
   \rowcolor{green!10}
  & \textbf{\Alg} &\textbf{64.75}  &\textbf{7.33}  &\textbf{69.43}  &\textbf{5.77}  &\textbf{62.12}  &\textbf{5.95}  &\textbf{65.33}  &\textbf{5.35}  &\textbf{70.77}  &\textbf{3.60}  &\textbf{61.59}  &\textbf{6.12} &\textbf{64.99} &\textbf{5.39} &68.12  &\textbf{4.53}\\
\hline
\end{tabular}
}
%\vspace{-0.6cm}
\label{tab:benchmark}
\end{table*}

\section{Experiments}
\textbf{Models and Datasets.}  
We evaluate our method on the full LLaMA model family, including LLaMA-1 (7B–30B)~\citep{touvron2023llama}, LLaMA-2 (7B–70B)~\citep{touvron2023llama}, and LLaMA-3 (8B, 70B). Perplexity (PPL) is measured on the WikiText2~\citep{merity2016pointer} test set. However, PPL alone does not fully reflect post-quantization performance, so we also report zero-shot accuracy on nine downstream tasks using the lm-evaluation-harness (v0.4.4)~\citep{gao2023framework}. These tasks include BoolQ~\citep{clark2019boolq}, HellaSwag~\citep{zellers2019hellaswag}, LAMBADA (OpenAI)~\citep{radford2019language}, OpenBookQA~\citep{mihaylov2018can}, PIQA~\citep{bisk2020piqa}, SIQA~\citep{sap2019socialiqa}, WinoGrande~\citep{sakaguchi2021winogrande}, ARC-Easy, and ARC-Challenge~\citep{boratko2018systematic}.

\noindent\textbf{Baselines.}  
We compare our method with standard RTN and several strong baselines, including SmoothQuant~\citep{xiao2023smoothquant}, GPTQ~\citep{frantar2022gptq}, Quarot~\citep{ashkboos2024quarot}, and SpinQuant~\citep{liu2024spinquant}, covering both weight-only and weight-activation quantization. All activations are quantized using per-token asymmetric quantization. Residual weights are quantized using GPTQ~\citep{frantar2022gptq}.

\noindent\textbf{Implementation Details.}
To perform smoothing, we follow SmoothQuant and compute a per-channel smoothing factor. The optimal migration strength $\alpha$ for each layer is selected offline by minimizing the mean squared error (MSE) of the layer outputs after frequency-domain truncation on a calibration set. We apply different low-frequency group counts based on the target bit width. For example, in 4-bit quantization, each channel retains 16 low-frequency groups. To ensure fair comparison, we control the bit width of residual weights such that the total bit width after compression is aligned with baseline methods. A set of 256 randomly sampled examples from WikiText2 is used for calibration.

% \noindent\textbf{Baselines and Implementation Details.}
% In addition to the basic RTN method, we benchmark our approach against SmoothQuant\cite{xiao2023smoothquant}, GPTQ\cite{frantar2022gptq}, and other state-of-the-art methods such as Quarot\cite{ashkboos2024quarot} and SpinQuant\cite{liu2024spinquant} for both weight-only and weight-activation quantization. All activations are quantized using per-token asymmetric quantization without any pruning. We use GPTQ\cite{frantar2022gptq} to quantize the residual weights.
% The smoothing factor $\lambda \in \mathbb{R}^m$ is a per-channel vector, with the $i$-th element computed as:
% $\lambda_i = \max(|X_{:,i}|)^\alpha / \max(|W_{:,i}|)^{1-\alpha}$
% This follows SmoothQuant\cite{xiao2023smoothquant}, where $X \in \mathbb{R}^{b \times m}$ and $W \in \mathbb{R}^{m \times n}$. The optimal migration strength $\alpha$ for each layer is determined offline by searching to minimize the mean squared error (MSE) of the layer outputs after frequency-domain truncation on the calibration dataset. For different quantization bit widths, we apply different low-frequency group counts in the frequency-domain truncation. For example, for 4-bit quantization, we set the truncation group count to 16 for each channel. To ensure fair comparison, we control the quantization bit width of the residual weights according to the truncation group count, so that the overall quantization bit width after compression matches that of the baseline methods.
% We use 256 randomly selected samples from WikiText2\cite{merity2016pointer} as the calibration dataset.

\begin{table*}[ht]
\centering
\caption{Prefill time and Memory usage of LLaMA models with different parameter sizes and sequence lengths, compared between our 4-bit implementation and FP16. All tests were conducted on a Transformer block with batch size 4 on a 3090 GPU.}
%\vspace{-0.3cm}
%\resizebox{\textwidth}{!}{
\tablestyle{5pt}{1.3}
\begin{tabular}{|c|c|ccc|ccc|}
\hline
\multirow{2}{*}{\textbf{Model}} & \multirow{2}{*}{\textbf{Seqlen}} & \multicolumn{2}{c}{\textbf{Prefill Time}}  & \multirow{2}{*}{\textbf{Prefill Speedup}} & \multicolumn{2}{c}{\textbf{Memory}} &\multirow{2}{*}{\textbf{Memory Saving}} \\
 & &\textbf{FP16} &\textbf{INT4} & &\textbf{FP16} &\textbf{INT4} & \\
\hline
\multirow{6}{*}{LLaMA2-7B} & 256 & 8.050ms & 3.579ms & 2.249x & 0.411GB & 0.132GB & 3.114x \\
 & 512 & 14.904ms & 6.759ms & 2.205x & 0.435GB & 0.138GB & 3.152x \\
 & 1024 & 27.989ms & 12.964ms & 2.159x & 0.483GB & 0.158GB & 3.057x \\
 & 2048 & 54.276ms & 26.208ms & 2.071x & 0.577GB & 0.212GB & 2.722x \\
 & 4096 & 112.230ms & 53.905ms & 2.082x & 0.766GB & 0.302GB & 2.536x \\
 & 8192 & 244.675ms & 122.460ms & 1.998x & 1.147GB & 0.495GB & 2.317x \\
\hline
\multirow{6}{*}{LLaMA3-8B} & 256 & 8.035ms & 3.510ms & 2.289x & 0.430GB & 0.126GB & 3.413x \\
 & 512 & 15.545ms & 6.663ms & 2.333x & 0.442GB & 0.134GB & 3.299x \\
 & 1024 & 29.169ms & 13.086ms & 2.229x & 0.466GB & 0.151GB & 3.086x \\
 & 2048 & 57.470ms & 26.312ms & 2.188x & 0.513GB & 0.188GB & 2.729x \\
 & 4096 & 117.523ms & 53.082ms & 2.214x & 0.608GB & 0.278GB & 2.187x \\
 & 8192 & 256.394ms & 119.198ms & 2.151x & 0.795GB & 0.400GB & 1.988x \\
\hline
\multirow{6}{*}{LLaMA2-13B} & 256 & 11.449ms & 4.667ms & 2.453x & 0.634GB & 0.178GB & 3.562x \\
 & 512 & 21.195ms & 8.602ms & 2.464x & 0.663GB & 0.192GB & 3.453x \\
 & 1024 & 41.752ms & 17.182ms & 2.430x & 0.723GB & 0.220GB & 3.286x \\
 & 2048 & 81.965ms & 34.864ms & 2.351x & 0.841GB & 0.283GB & 2.972x \\
 & 4096 & 199.046ms & 81.710ms & 2.436x & 1.079GB & 0.404GB & 2.671x \\
 & 8192 & 359.409ms & 162.335ms & 2.214x & 1.551GB & 0.642GB & 2.416x \\
\hline
\multirow{6}{*}{LLaMA-30B} & 256 & 18.682ms & 6.485ms & 2.881x & 1.047GB & 0.285GB & 3.674x \\
 & 512 & 34.393ms & 12.743ms & 2.699x & 1.085GB & 0.305GB & 3.557x \\
 & 1024 & 66.880ms & 24.835ms & 2.693x & 1.162GB & 0.343GB & 3.388x \\
 & 2048 & 157.500ms & 59.886ms & 2.630x & 1.315GB & 0.422GB & 3.116x \\
 & 4096 & 272.355ms & 105.523ms & 2.581x & 1.625GB & 0.577GB & 2.816x \\
 & 8192 & 576.555ms & 234.086ms & 2.463x & 2.242GB & 0.889GB & 2.522x \\
\hline
\end{tabular}
%}
\label{tab:speedup}
%\vspace{-0.4cm}
\end{table*}

\begin{table*}[t]
\centering
\caption{Ablation study on the impact of different methods on WikiText2 perplexity (PPL) and zero-shot$^9$ accuracy for LLaMA-7B, LLaMA2-7B, and LLaMA3-8B.}
%\vspace{-0.3cm}
%\resizebox{0.48\textwidth}{!}{%
\tablestyle{4pt}{1.3}
\begin{tabular}{ccc|cc|cc|cc}
\toprule
\multicolumn{3}{c|}{\textbf{Method}} & \multicolumn{2}{c|}{\textbf{LLaMA-7B}} & \multicolumn{2}{c|}{\textbf{LLaMA2-7B}} & \multicolumn{2}{c}{\textbf{LLaMA3-8B}}\\
\midrule
\textbf{Quant} & \textbf{Smooth} & \textbf{Trunc.}
 & \textbf{Wiki $\downarrow$} & \textbf{0-shot$^9$ $\uparrow$} 
 & \textbf{Wiki $\downarrow$} & \textbf{0-shot$^9$ $\uparrow$}   & \textbf{Wiki $\downarrow$} & \textbf{0-shot$^9$ $\uparrow$} \\
\midrule
\checkmark & & &9e3  &25.34  & nan & 26.44 &8e3 &24.42 \\
 \checkmark & \checkmark & & 3e2 & 34.42 & nan & 32.13 &1e3 &33.04 \\
  \checkmark &  &\checkmark & 24.57 & 54.72 & 26.79 & 52.88 &27.75 &55.08 \\
 \rowcolor{green!10}
  \checkmark & \checkmark & \checkmark &\textbf{6.05} &\textbf{61.85} &\textbf{5.88} &\textbf{62.88} &\textbf{7.25} &\textbf{64.75} \\
\bottomrule
\end{tabular}
%}
%\vspace{-0.3cm}
\label{tab:ablation}
\end{table*}

\subsection{Overall Results}
\textbf{Quantization Performance.}
As shown in Table~\ref{tab:benchmark}, \Alg consistently outperforms prior SOTA methods across a wide range of models and quantization settings. In the 4-16-16 setting, \Alg maintains over 99\% of full-precision (FP) accuracy on zero-shot tasks, surpassing both activation-aware and weight-only baselines. Compared to weight-only approaches such as GPTQ and AWQ, \Alg achieves a narrower gap to FP performance, particularly on challenging models like LLaMA-3-8B, where it incurs only a 1.21\% accuracy drop---significantly lower than the $>1.55\%$ degradation seen in competing methods.
Under the 4-4-16 quantization setting, which imposes stricter constraints on both weights and activations, \Alg consistently outperforms SpinQuant by over 1 percentage point in accuracy across multiple benchmark models. Even under the extreme 4-4-4 quantization setting, \Alg achieves notable accuracy improvements, demonstrating strong robustness to aggressive compression. These results validate the effectiveness of our frequency-domain approach. Specifically, the low-frequency truncation branch outperforms prior methods based solely on rotation or smoothing by better suppressing activation outliers and addressing distributional imbalances, thereby enhancing quantization stability.

\begin{table*}[t]
\centering
\caption{Impact of the number of truncation groups in the low-frequency truncation branch on WikiText perplexity (PPL) and zero-shot$^9$ accuracy for LLaMA models.}
%\vspace{-0.3cm}
%\resizebox{0.48\textwidth}{!}{%
\tablestyle{3pt}{1.3}
\begin{tabular}{c|c|c|cc|cc|cc}
\toprule
{\textbf{Trunc.}} & {\textbf{Model Size}} &{\textbf{Latency}} & \multicolumn{2}{c|}{\textbf{LLaMA-7B}} & \multicolumn{2}{c}{\textbf{LLaMA2-7B}} & \multicolumn{2}{c}{\textbf{LLaMA3-8B}}\\

 \textbf{Groups} &\textbf{Overhead} & \textbf{Overhead} &\textbf{Wiki$\downarrow$} & \textbf{0-shot$^9$ $\uparrow$} 
 & \textbf{Wiki$\downarrow$} & \textbf{0-shot$^9$ $\uparrow$}   & \textbf{Wiki $\downarrow$} & \textbf{0-shot$^9$ $\uparrow$} \\
\midrule
16 &2.7\% &5.2\%  &6.04  &61.89  & 5.87 & 62.91 &7.24 &64.78 \\
32 &5.5\% &7.3\%  &6.03 & 62.01 & 5.85 & 63.11 &7.21 &65.12 \\
64 &11.2\%  &12.1\% &5.99 & 62.88 & 5.81 & 64.09 &7.08 &66.03 \\

\bottomrule
\end{tabular}
%}
%\vspace{-0.6cm}
\label{tab:tradeoff}
\end{table*}

\begin{table}[ht]
\centering
% \vspace{-0.3cm}
\caption{Comparison of Importance Metrics for Compression Allocation in \Alg under a 20\% compression ratio on LLaMA-7B and LLaMA2-7B.}
%\vspace{-0.3cm}
\tablestyle{3pt}{1.3}
\footnotesize
\begin{tabular}{c|cc|cc}
\toprule
\multirow{2}{*}{Importance Metrics} & \multicolumn{2}{c|}{LLaMA-7B} & \multicolumn{2}{c}{LLaMA2-7B} \\

& WikiText2 $\downarrow$ & PTB $\downarrow$ 
& WikiText2 $\downarrow$ & PTB $\downarrow$  \\
\midrule
Original &6.60 &66.00 &6.33 &33.63 \\
Abs Mean &6.58 &56.60 &6.32 &33.22 \\
Abs Max &6.72 &53.67 &6.45 &34.17 \\
L2 Norm &6.59 &55.19 &6.31 &33.08 \\
\rowcolor{green!10}
\textbf{Spectral Entropy} &\textbf{6.55} &\textbf{47.52} &\textbf{6.30} &\textbf{32.64} \\
\bottomrule
\end{tabular}
\label{tab:metrics}
%\vspace{-0.6cm}
\end{table}

\noindent\textbf{Speedup and Memory Savings.}
\Alg achieves 4-bit quantization with negligible accuracy loss, enabling practical low-bit inference. To assess its efficiency, we follow the measurement setup used in prior work \cite{hu2025ostquant}, and evaluate the LLaMA series on an NVIDIA 3090 GPU. As shown in Table~\ref{tab:speedup}, \Alg delivers over 2× speedup across models, reaching nearly 2.5× on the challenging LLaMA-30B, due to efficient low-precision computation and reduced memory access overhead. Additionally, \Alg achieves over 3× memory savings on average. The added cost from the low-frequency truncation branch is minimal, introducing no significant computational overhead.

\subsection{Ablation Study}
To validate the contribution of each component in our method, we conducted detailed ablation experiments. As shown in Table~\ref{tab:ablation}, direct W4A4 quantization leads to a notable drop in performance compared to the full-precision model. Introducing the Smooth operation prior to quantization yields modest gains, but the improvement is limited. Adding a low-frequency truncation branch alone also fails to deliver satisfactory results. In contrast, our full method significantly improves quantization performance by first migrating activation outliers into the weight parameters via smoothing, and then quantizing the residual weights after decomposition. The low-frequency truncation branch effectively absorbs the migrated outliers, resulting in a robust low-bit model.

\subsection{Trade-off of Increasing Truncation Groups}
Table~\ref{tab:tradeoff} summarizes the impact of varying the number of truncation groups in \Alg's low-frequency branch. As the number of groups increases, quantization performance improves consistently, approaching full-precision accuracy. However, this gain comes with additional parameter storage and higher inference latency. To strike a balance between accuracy and efficiency, we adopt a configuration with 16 truncation groups. Furthermore, by applying mixed-precision quantization to the residual weights, \Alg maintains the same parameter count as baseline methods while delivering superior performance.

\subsection{Comparison of Importance Metrics for Truncation}
\label{sec:metrics}

We compare four channel-wise importance metrics for allocating compression budgets under a fixed 20\% compression ratio: Mean Absolute Value (Abs Mean), Maximum Absolute Value (Abs Max), L2 Norm, and our proposed spectral entropy. These metrics, commonly used in quantization and pruning, capture different aspects of channel salience. As shown in Table~\ref{tab:metrics}, spectral entropy consistently yields the lowest perplexity across all benchmarks and models, validating its robustness and effectiveness. Unlike magnitude-based metrics, it captures the energy distribution across frequency components, enabling better identification of structurally important channels. These findings empirically support our frequency-aware design and highlight the advantage of spectral-domain importance estimation in compression-aware allocation.

\section{Conclusion}
In this paper, we propose SpecQuant, a novel frequency-domain quantization method for ultra-low-bit LLMs, which effectively addresses the long-standing challenge of outlier management through spectral-domain processing. \Alg combines activation smoothing with channel-adaptive Fourier decomposition, ensuring that amplified weights in the frequency domain do not impact the low-frequency components, which retain most of the signal energy. Experiments across multiple zero-shot tasks validate the effectiveness of our method across various network architectures.

%\clearpage

\bibliography{aaai2026}

\clearpage
\section{Appendix}
\label{sec:appendix}

\subsection{Pseudocode of \Alg}
\label{sec:A}
As shown in Algorithm \ref{alg:method}, we propose a Low-Frequency Fourier Projection quantization framework (\Alg) for efficient compression of weight matrices and activation tensors in large-scale language models. The core innovation of our method lies in the introduction of an auxiliary low-frequency spectral truncation branch, which effectively mitigates channel-wise quantization errors inherent in both weight matrices and activation tensors by performing compression in the frequency domain. The process starts by smoothing the input activation and weight matrices to reduce outliers, followed by a Low-Frequency Fourier Projection step. During this step, each channel's smoothed weight vector is transformed into the frequency domain using the Fast Fourier Transform (FFT), where low-frequency components are retained and high-frequency noise is suppressed. The truncated frequency components are then reconstructed to form the compressed weight vector. In the quantization process, the low-frequency components are quantized with higher precision (e.g., 16-bit), while the residuals are quantized with lower precision (e.g., 4-bit). This strategy ensures that the majority of useful information is preserved, while memory usage is significantly reduced. The final output is the quantized model, consisting of the compressed activation and weight matrices, offering a storage-efficient solution for large language models while maintaining high performance during inference.

\begin{algorithm}[htbp]
\caption{Quantization Framework}
\label{alg:method}
\begin{algorithmic}[1] % [1] 表示显示行号
\REQUIRE $\mathbf{X} \in \mathbb{R}^{N \times M}$: Input activation matrix, $\mathbf{W} \in \mathbb{R}^{N \times D}$: Weight matrix, $\boldsymbol{\lambda} \in \mathbb{R}^N$: Per-channel smoothing factor, $\rho \in [0, 1]$: Compression ratio
\ENSURE $\hat{\mathbf{X}} \hat{\mathbf{W}}$: Quantized model output

\STATE \textbf{Step 1: Smoothing Preprocessing}
\FOR{each channel $j \in \{1, 2, \dots, N\}$}
    \STATE $\hat{\mathbf{X}}_j \gets \mathbf{X}_j \cdot \text{diag}(\boldsymbol{\lambda})^{-1}$ \COMMENT{Smoothed activation for channel $j$}
    \STATE $\hat{\mathbf{W}}_j \gets \mathbf{W}_j \cdot \text{diag}(\boldsymbol{\lambda})$ \COMMENT{Smoothed weight for channel $j$}
\ENDFOR

\STATE \textbf{Step 2: Low-Frequency Fourier Projection for Weight Vectors}
\FOR{each channel $j \in \{1, 2, \dots, N\}$}
    \STATE $\mathbf{W}_{\text{freq}, j} \gets \mathcal{F}(\hat{\mathbf{W}}_j)$ \COMMENT{Apply FFT to weight vector}
    \STATE $\bar{a}_j \gets -\sum_f p(f) \log p(f)$ \COMMENT{Compute spectral entropy}
    \STATE $\rho_j \gets \frac{\exp(\alpha \bar{a}_j)}{\sum_{l=1}^{D} \exp(\alpha \bar{a}_l)}$ \COMMENT{Truncation ratio based on spectral entropy}
    \STATE $k_j \gets \lfloor \rho_j \cdot N \rfloor$ \COMMENT{Number of frequency components to retain}
    \STATE $\mathbf{W}_{\text{freq}, j, \text{truncated}} \gets \mathbf{W}_{\text{freq}, j}[:k_j]$ \COMMENT{Retain top-$k_j$ low-frequency components}
    \STATE $\hat{\mathbf{W}}_j' \gets \mathcal{F}^{-1}(\mathbf{W}_{\text{freq}, j, \text{truncated}})$ \COMMENT{Reconstruct weight vector via inverse FFT}
\ENDFOR

\STATE \textbf{Step 3: Quantization Process}
\FOR{each channel $j \in \{1, 2, \dots, N\}$}
    \STATE $\hat{\mathbf{X}} \mathbf{W}' \gets \hat{\mathbf{X}} \cdot \hat{\mathbf{W}}'$ \COMMENT{Matrix multiplication with compressed weights}
    \STATE $\mathbf{R}_j \gets \hat{\mathbf{W}}_j - \hat{\mathbf{W}}_j'$ \COMMENT{Compute residual}
    \STATE $Q(\mathbf{R}_j) \gets \text{Quantize}(\mathbf{R}_j)$ \COMMENT{Quantize residual (e.g., 4-bit)}
    \STATE $Q(\hat{\mathbf{W}}_j') \gets \text{Quantize}(\hat{\mathbf{W}}_j')$ \COMMENT{Quantize low-frequency components (e.g., 16-bit)}
    \STATE $\hat{\mathbf{X}} \hat{\mathbf{W}}_j \gets Q(\hat{\mathbf{X}} \mathbf{W}') + Q(\mathbf{R}_j)$ \COMMENT{Combine quantized results}
\ENDFOR

\STATE \textbf{Step 4: Output Quantized Model}
\RETURN $\hat{\mathbf{X}} \hat{\mathbf{W}}$ \COMMENT{Return final quantized activation and weight matrices}
\end{algorithmic}
\end{algorithm}

\subsection{Compression Error Analysis of Spectrum Compression with SVD}
\label{sec:with SVD}
Let $\mathbf{W} \in \mathbb{R}^{C_{\text{in}} \times C_{\text{out}}}$ denote a weight matrix in a deep neural network. Empirical studies have shown that such weight matrices often exhibit strong channel-wise spectral decay, meaning that most of the energy of each channel $\mathbf{W}[:,j]$ is concentrated in its low-frequency components. Under this assumption, we prove that, given a fixed total parameter budget $B$, our proposed channel-wise frequency-domain compression achieves strictly smaller reconstruction error than truncated SVD in terms of the Frobenius norm.

To ensure fair comparison, we first match the total parameter budgets. Truncated SVD retains $k$ singular triplets, consuming $B_{\text{SVD}} = k(C_{\text{in}} + C_{\text{out}} + 1)$ parameters. In contrast, our spectral compression retains $k_j$ frequency components for each channel, consuming $B_{\text{spectral}} = 2 \sum_{j=1}^{C_{\text{out}}} k_j$ parameters. Setting $B_{\text{SVD}} = B_{\text{spectral}} = B$, we guarantee that both methods are constrained under the same storage budget.

Next, we formulate the reconstruction errors. For SVD, the loss is given by
\begin{equation}
\mathcal{L}_{\text{SVD}} = \sum_{i=k+1}^{\min(C_{\text{in}}, C_{\text{out}})} \sigma_i^2
\end{equation}
where $\sigma_i$ are singular values sorted in descending order.

For spectral compression, by Parseval’s theorem, the loss per channel is
\begin{equation}
\|\mathbf{W}[:,j] - \hat{\mathbf{W}}_{\text{spectral}}[:,j]\|_2^2 = \sum_{f=k_j}^{C_{\text{in}}-1} |\mathbf{W}_{\text{freq}}[f,j]|^2
\end{equation}
and the total loss sums across channels as
\begin{equation}
\mathcal{L}_{\text{Freq}} = \sum_{j=1}^{C_{\text{out}}} \sum_{f=k_j}^{C_{\text{in}}-1} |\mathbf{W}_{\text{freq}}[f,j]|^2
\end{equation}

Given the assumed spectral decay
\begin{equation}
|\mathbf{W}_{\text{freq}}[f,j]| \leq \frac{C_j}{f^r} \quad (r > 1),
\end{equation}
we can upper-bound the loss for each channel:
\begin{equation}
\sum_{f=k_j}^{\infty} |\mathbf{W}_{\text{freq}}[f,j]|^2 \leq \frac{C_j^2}{(2r-1)(k_j-1)^{2r-1}}
\end{equation}
where the integral approximation $\sum \frac{1}{f^{2r}} \leq \int \frac{dx}{x^{2r}}$ is used.

Summing across all channels, the total spectral compression loss satisfies:
\begin{equation}
\mathcal{L}_{\text{Freq}} \leq \sum_{j=1}^{C_{\text{out}}} \frac{C_j^2}{(2r-1)(k_j-1)^{2r-1}}
\end{equation}

By contrast, the SVD loss $\mathcal{L}_{\text{SVD}}$ depends on the decay of singular values, which represent global variations across both input and output dimensions. In practice, due to complex feature mixing and lack of global low-rank structures in intermediate layers of neural networks, the singular values often decay much slower than individual channel spectral coefficients.

Consequently, while spectral compression exploits localized channel smoothness to retain dominant energy very efficiently, truncated SVD, constrained by global low-rankness, loses more useful energy when aggressively compressed under the same budget. Thus, we conclude:
\begin{equation}
\|\mathbf{W} - \hat{\mathbf{W}}_{\text{spectral}}\|_F < \|\mathbf{W} - \hat{\mathbf{W}}_{\text{SVD}}\|_F
\end{equation}
which proves that our method achieves strictly lower reconstruction error under equivalent compression ratios.

\subsection{Comparison with SVDQuant}
\label{sec:with SVDQuant}
\begin{figure*}[h]
    \centering
    \includegraphics[width = \textwidth]{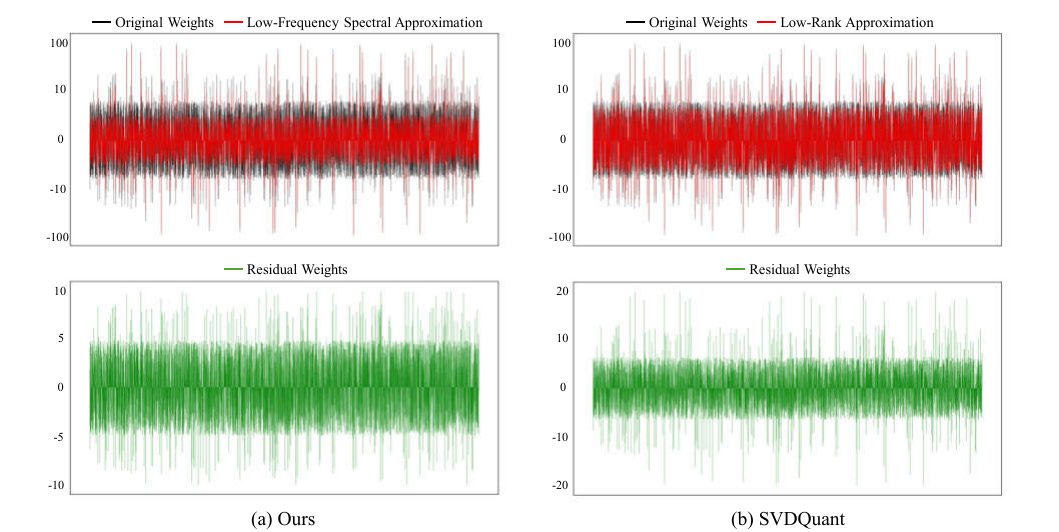}
    \caption{Comparison of weights decomposed by \Alg and SVDQuant.}
    \label{fig:SVDQuant}
\end{figure*}

In this section, we conduct a comparative analysis between our method and SVDQuant\cite{li2024svdqunat}. While SVDQuant employs singular value decomposition (SVD)-based low-rank approximation for outlier absorption, our approach introduces a novel Channel-wise Low-Frequency Spectral Approximation mechanism. As visually demonstrated in Figure \ref{fig:SVDQuant}, the global matrix decomposition strategy of conventional SVD methods shows inherent limitations when handling cross-channel distributed outliers: although Figure \ref{fig:SVDQuant}(b) indicates partial outlier absorption by SVD, residual outliers persist in cases of significant cross-channel variations (evident from remaining anomalies in the residual matrix). In contrast, our channel-wise spectral approximation method exhibits two key advantages: (1) It preserves channel locality characteristics, overcoming the deficiencies of traditional low-rank methods in cross-channel outlier handling; (2) It innovatively exploits the inherent smoothness of weight distributions in the frequency domain. Figure \ref{fig:SVDQuant}(a) demonstrates that our method effectively eliminates outliers in residual matrices, significantly enhancing quantization compatibility.

% \subsection{Comparison of Importance Metrics for Truncation}
% \label{sec:metrics}
% \begin{table*}[ht]
% \centering
% \caption{Comparison of Importance Metrics for Compression Allocation in \Alg under a 20\% compression ratio on LLaMA-7B and LLaMA2-7B.}

% \begin{tabular}{c|cc|cc}
% \toprule
% \multirow{2}{*}{Importance Metrics} & \multicolumn{2}{c|}{LLaMA-7B} & \multicolumn{2}{c}{LLaMA2-7B} \\

% & WikiText2 $\downarrow$ & PTB $\downarrow$ 
% & WikiText2 $\downarrow$ & PTB $\downarrow$  \\
% \midrule
% \Alg (original) &6.60 &66.00 &6.33 &33.63 \\
% \Alg (Abs Mean) &6.58 &56.60 &6.32 &33.22 \\
% \Alg (Abs Max) &6.72 &53.67 &6.45 &34.17 \\
% \Alg (L2 Norm) &6.59 &55.19 &6.31 &33.08 \\
% \rowcolor{green!10}
% \textbf{\Alg (Spectral Entropy)} &\textbf{6.55} &\textbf{47.52} &\textbf{6.30} &\textbf{32.64} \\
% \bottomrule
% \end{tabular}
% \label{tab:metrics}
% \end{table*}

\subsection{Visualization Comparison of \Alg and SVD Compression Distributions}
\label{sec:visualization}
\begin{figure*}[h]
    \centering
    \includegraphics[width=1\textwidth]{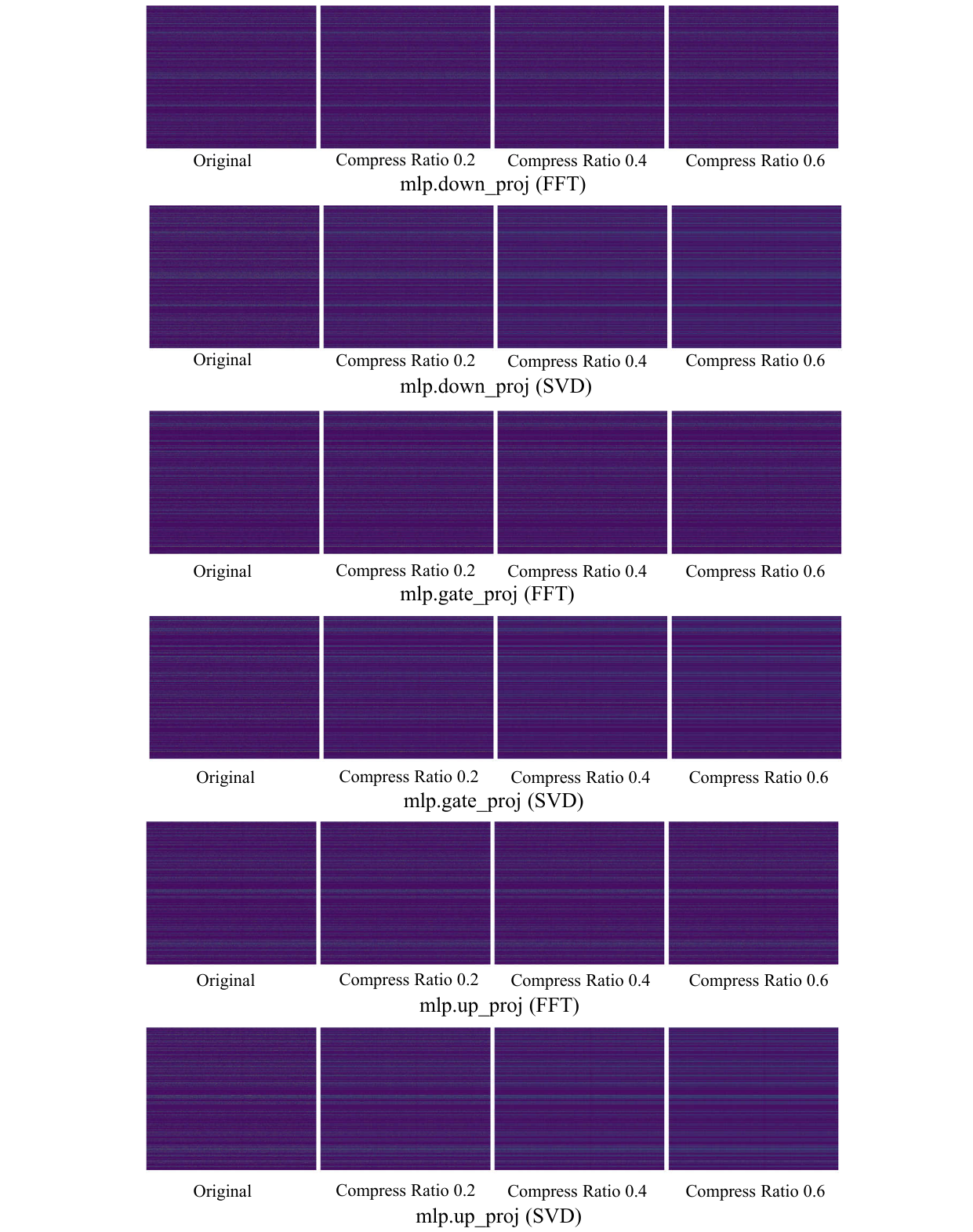}
    \caption{Distribution comparison of the original weight magnitudes and the approximated weights by \Alg and SVD at different compression ratios for the first fully connected layer of the LLaMA-2 7B model.}
    \label{fig:fig1}
\end{figure*}

\begin{figure*}[h]
    \centering
    \includegraphics[width=1\textwidth]{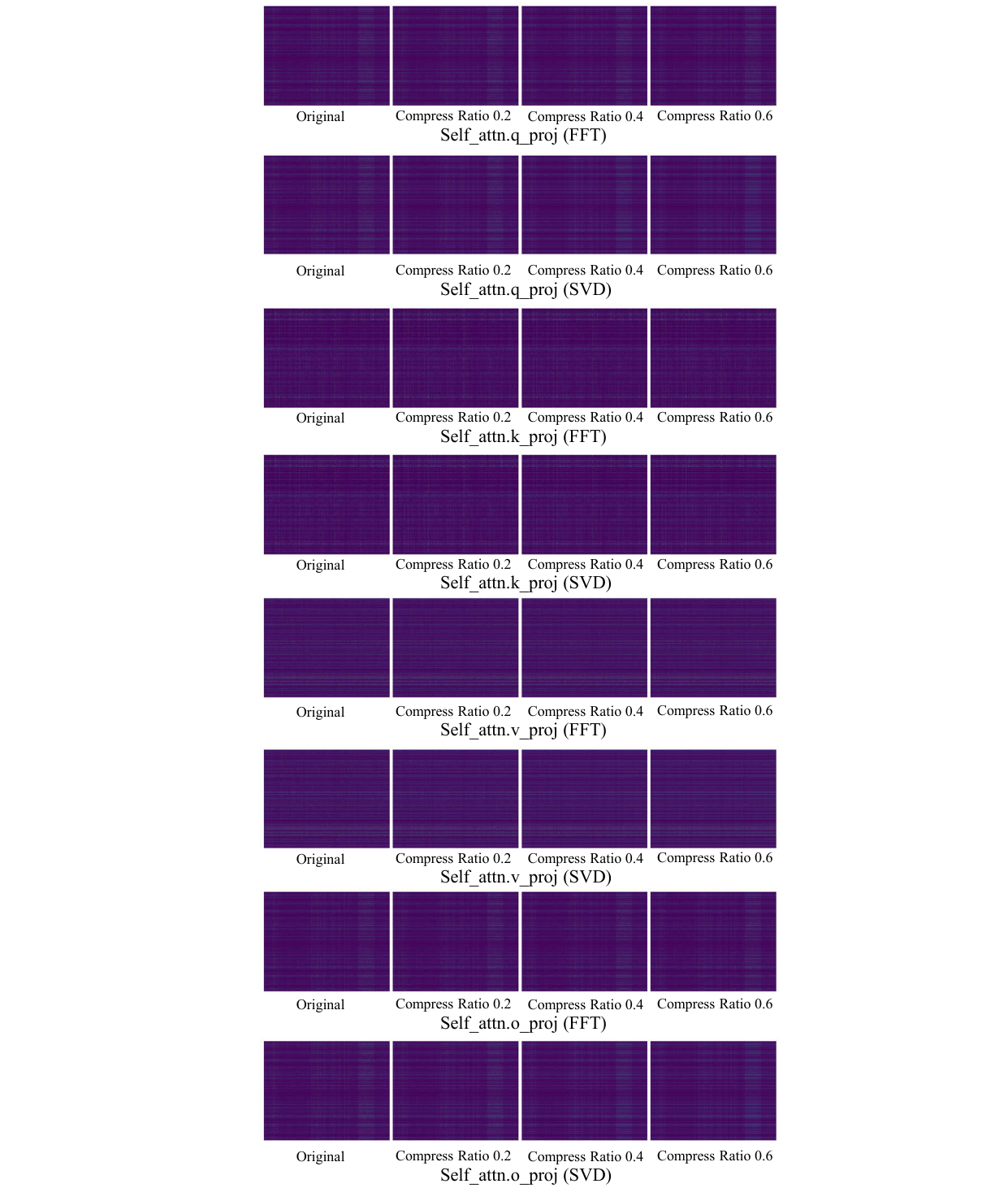}
    \caption{Distribution comparison of the original weight magnitudes and the approximated weights by \Alg and SVD at different compression ratios for the first self-attention layer of the LLaMA-2 7B model.}
    \label{fig:fig2}
\end{figure*}

Figures~\ref{fig:fig1} and~\ref{fig:fig2} provide a comparative visualization of transposed weight matrices from self-attention and fully connected layers, compressed at the same ratio using our method (\Alg) and the standard SVD approach. Two key observations can be drawn.
First, the weight matrices exhibit strong structural patterns along the channel dimension, with certain channels consistently demonstrating greater representational importance. This observation supports the design of our channel-wise compression strategy.
Second, under high compression ratios, SVD-reconstructed weights show significant parameter polarization—manifested as over-amplification in high-norm channels and suppression of near-zero weights—disrupting the original structure. In contrast, \Alg better preserves the local self-similarity of the weight distribution, even under aggressive compression, resulting in reconstructions that remain much closer to the uncompressed baseline.
These findings underscore the natural compatibility of frequency-domain compression with sparse representations, and provide a theoretical basis for the superior generalization performance observed in \Alg-compressed models.

\end{document}